\documentclass{article}
\pdfoutput=1

\usepackage{microtype}
\usepackage{graphicx}
\usepackage{subfigure}
\usepackage{booktabs} 

\usepackage{hyperref}

\usepackage[accepted]{icml2024}

\usepackage[amsthm]{ntheorem}
\usepackage{amsmath}
\usepackage{amssymb}
\usepackage{mathtools}
\usepackage{xspace}

\usepackage[capitalize,noabbrev]{cleveref}

\theoremstyle{plain}

\theoremstyle{definition}

\theoremstyle{remark}

\newcommand{\ie}{{\textit{i.e., }\xspace}}

\usepackage{amsfonts}
\usepackage{enumitem}
\usepackage{comment}
\usepackage{mdframed}
\usepackage{wrapfig}
\usepackage{makecell}

\usepackage{tcolorbox}
\tcbuselibrary{skins} 
\newtcolorbox{dialogbox}{
    enhanced,
    boxrule=1pt, 
    colback=black!10, 
    colframe=black, 
    left=1pt, 
    right=1pt, 
    top=3pt, 
    bottom=3pt, 
}

\theoremstyle{nonumberplain}

\newmdtheoremenv[%
  backgroundcolor=white,
  linecolor=red!60!black,
  linewidth=2pt,
  topline=true,
  rightline=false,
  skipabove=10pt,
  skipbelow=10pt,
  leftline=false]{ouresponse}{}

\theoremstyle{nonumberplain}
\newmdtheoremenv[%
  backgroundcolor=gray!40,
  linecolor=red!50!black,
  linewidth=2pt,
  topline=false,
  rightline=false,
  skipabove=10pt,
  skipbelow=10pt,
  leftline=false]{ourbox}{}

\usepackage[textsize=tiny]{todonotes}

\icmltitlerunning{Faithfulness vs. Plausibility: On the (Un)Reliability of Explanations from Large Language Models}
\begin{document}

\twocolumn[
\icmltitle{Faithfulness vs. Plausibility: On the (Un)Reliability of Explanations \\from Large Language Models}



\icmlsetsymbol{equal}{*}

\begin{icmlauthorlist}
\icmlauthor{Chirag Agarwal}{yyy}
\icmlauthor{Sree Harsha Tanneru}{yyy}
\icmlauthor{Himabindu Lakkaraju}{yyy}
\end{icmlauthorlist}

\icmlaffiliation{yyy}{Harvard University, Cambridge, MA, USA}

\icmlcorrespondingauthor{Chirag Agarwal}{cagarwal@hbs.edu}

\icmlkeywords{Machine Learning}

\vskip 0.3in
]



\printAffiliationsAndNotice{}  

\begin{abstract}
\looseness=-1 Large Language Models (LLMs) are deployed as powerful tools for several natural language processing (NLP) applications. Recent works show that modern LLMs can generate self-explanations (SEs), which elicit their intermediate reasoning steps for explaining their behavior. Self-explanations have seen widespread adoption owing to their conversational and plausible nature. However, there is little to no understanding of their faithfulness. In this work, we discuss the dichotomy between faithfulness and plausibility in SEs generated by LLMs. We argue that while LLMs are adept at generating plausible explanations -- seemingly logical and coherent to human users -- these explanations do not necessarily align with the reasoning processes of the LLMs, raising concerns about their faithfulness. We highlight that the current trend towards increasing the plausibility of explanations, primarily driven by the demand for user-friendly interfaces, may come at the cost of diminishing their faithfulness. We assert that the faithfulness of explanations is critical in LLMs employed for high-stakes decision-making. Moreover, we emphasize the need for a systematic characterization of faithfulness-plausibility requirements of different real-world applications and ensure explanations meet those needs. While there are several approaches to improving plausibility, improving faithfulness is an open challenge. We call upon the community to develop novel methods to enhance the faithfulness of self explanations thereby enabling transparent deployment of LLMs in diverse high-stakes settings.\\
\textcolor{purple}{Content Warning: This paper may contain some offensive responses generated by LLMs.}

\end{abstract}

\section{Introduction}
\label{sec:intro}

\looseness=-1 In recent years, the advent of Large Language Models (LLMs) has revolutionized the field of natural language processing, offering impressive capabilities in generating human-like text~\citep{kaddour2023challenges}. However, LLMs are complex large-scale models trained on broad datasets and large scale compute, where their decision-making processes are not completely understood and remain an important bottleneck for deploying them to high-stakes applications.

Existing works show that these models, known for their extensive training on diverse datasets, demonstrate remarkable proficiency in generating self-explanations that are not only coherent but also contextually adaptable~\citep{brown2020language,wei2023chainofthought}, critical for
explaining their predictions in applications like health, commerce, and law. To this end, there has been an increasing focus on understanding the nature of the explanations generated by these LLMs. The concept of self-explanations extends beyond mere response generation; it encompasses the model's ability to explain its decisions in a manner understandable to humans. This capability is pivotal in applications where decision-making transparency is crucial, such as healthcare diagnostics, legal advice, and financial forecasting.

Despite their emergent capabilities~\citep{wei2022emergent}, LLMs face significant challenges in generating self-explanations, where one of the primary concerns is the gap between \textit{plausibility} and \textit{faithfulness}. While LLMs can generate responses that seem logical and coherent (\ie being plausible), these explanations may not accurately reflect the model's actual reasoning process (\ie being faithful)~\citep{Jacovi2020TowardsFI,Wiegreffe2021TeachMT}. This discrepancy raises questions about the reliability and trustworthiness of the model's outputs, especially in high-stakes scenarios.

Plausibility in self-explanations refers to how convincing and logical the explanations appear to humans. Given their extensive training, LLMs are adept at formulating explanations that align with human reasoning. However, these plausible explanations might be misleading if they do not correspond to the LLM's internal decision-making process. On the other hand, faithfulness represents the accuracy of explanations in illustrating the LLM's actual reasoning, \ie why and how the model reached a particular decision. We highlight the challenge of ensuring that self-explanations are both plausible and faithful, which is not always straightforward due to the complexity and opacity of LLMs and the dichotomy between these properties for specific downstream applications. For instance, in healthcare, an incorrect diagnostic explanation from an LLM could lead to severe implications for patient care. Similarly, in legal contexts, reliance on inaccurate explanations could result in erroneous legal advice. Hence, ensuring the faithfulness of explanations is as critical as their plausibility. Addressing the aforementioned challenges necessitates future research focused on simultaneously enhancing the plausibility and faithfulness of self-explanations, including the development of novel methodologies for better understanding the decision-making processes of LLMs and creating frameworks that can assess and ensure the faithfulness of explanations. Further, there is a need for developing evaluation metrics and benchmarks that can effectively measure the quality of self-explanations in terms of plausibility and faithfulness. 

Here, we review self-explanations generated by LLMs and some key connections between their plausibility and faithfulness properties. While there are many NLP applications of LLMs, we focus on high-stakes applications that require LLM-generated explanations to be plausible and/or faithful. We begin with an overview of self-explanations to highlight techniques like token importance and chain-of-thought reasoning relevant to this review. Next, we formally define and discuss the plausibility and faithfulness aspects of self-explanations and some challenges in achieving them. We underscore the potential of studying the implication of these properties in high-stakes applications, emphasizing when we need an explanation to be plausible or faithful and how this transparency in generated self-explanations is application-centric. Finally, we provide some open challenges and valuable insights for future research, contributing to a better understanding of the plausibility and faithfulness facet of explanations in LLMs.

\section{Self-Explanations}
\label{sec:se}

\begin{figure*}
    \centering
    \includegraphics[width=0.87\textwidth]{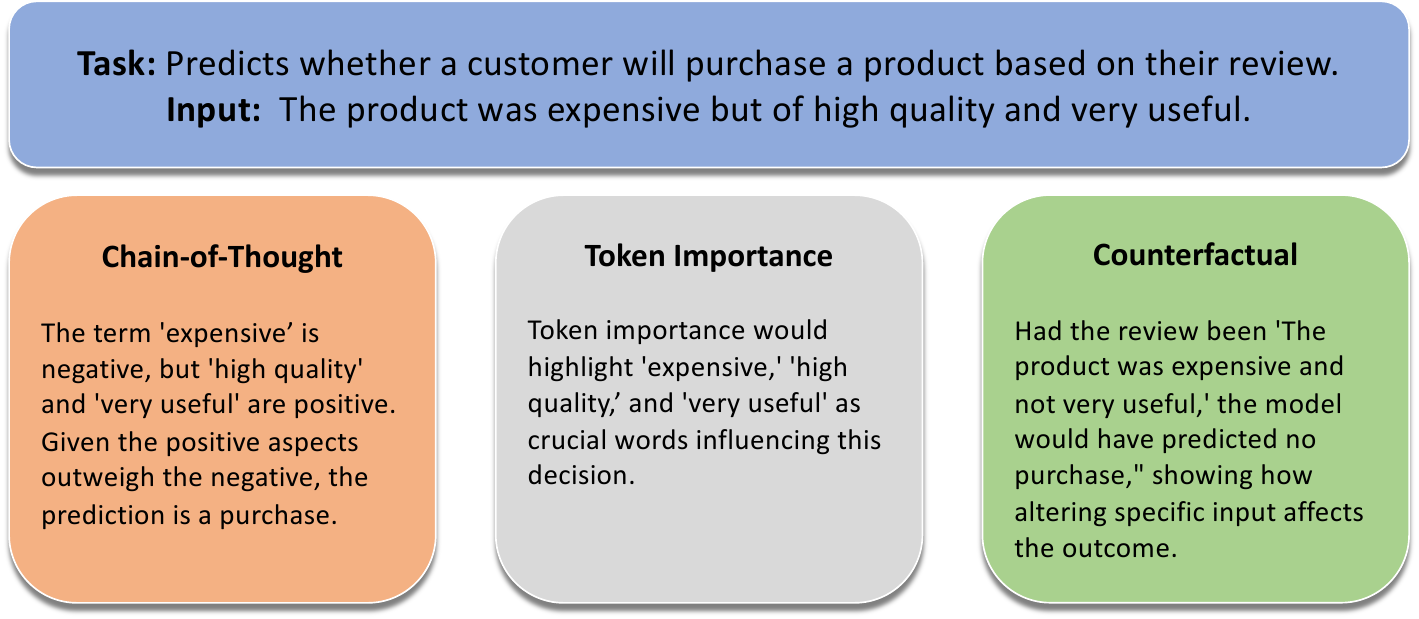}
    \vspace{-0.12in}
    \caption{\textbf{Self-explanations in LLMs.} Illustration of SE techniques for predictive modeling in customer review analysis. The figure shows three distinct SE methods: \textcolor{orange}{\textbf{Chain-of-Thought}}, which provides a logical reasoning path leading to the model's prediction; \textcolor{gray!90!black}{\textbf{Token Importance}}, which highlights key terms in the input that significantly influence the prediction; and \textcolor{green!50!black}{\textbf{Counterfactual}}, which demonstrates how modifications to the input could change the predicted outcome.}
    \vspace{-0.12in}
    \label{fig:nle}
\end{figure*}

Self-explanations (SEs) are a class of explanation methods that are generated by models to elicit the reasoning behind their decisions in human-understandable language. SEs are becoming increasingly crucial in the era of large language models, particularly for enhancing the trustworthiness of their generated responses. SEs can take various forms, including chain-of-thought reasoning, token importance, and counterfactual explanations, offering unique insights into the model's decision-making process (see Fig.~\ref{fig:nle}).

\looseness=-1 Chain-of-thought (CoT)~\citep{wei2023chainofthought} reasoning in SEs involves generating a sequence of intermediate thoughts or steps that lead to the final decision or response of an LLM. CoTs claim to elicit reasoning in LLMs and have shown noticeable performance gains in commonsense and math reasoning tasks. It is invaluable for users as they get a window into the thought process of these billion parameter LLMs and allows them to understand how they reach their decision. For instance, consider an LLM used to solve a math word problem: ``\textit{If John has 5 apples and gives 2 to Jane, how many does he have left?}'' the LLM doesn't just output ``3.'' Instead, it explains its reasoning steps as  ``\textit{1) John initially has 5 apples. 2) He gives 2 to Jane, so 5 - 2 = 3. 3) Therefore, John has 3 apples left.}'' This CoT reasoning makes the LLM's process more transparent and easier to trust.

Token importance~\citep{Li2015VisualizingAU,Wu2020PerturbedMP} in SEs involves highlighting specific input tokens (words or phrases) that significantly influence the model’s decision. By identifying these key tokens, users can understand which parts of the input had the most impact on the outcome. For instance, consider an LLM used to classify sentiments into positive and negative categories. When prompted to output the sentiment of an input, ``\textit{The movie was boring but visually ok,}'' the model might highlight ``\textit{boring}'' as an important token influencing the LLM negative sentiment analysis. This token-level importance allows users to understand which aspects of the input swayed the LLM's judgment.

Counterfactual explanations in SEs provide insights into how different inputs might lead to different LLM responses. These explanations answer ``\textit{what-if}'' scenarios, helping users understand how changes in the input could alter the model's decision. For instance, in the aforementioned example of sentiment analysis, a counterfactual explanation might be: ``\textit{If the word `boring' in the sentence `The movie was boring but visually ok,' were replaced with `great,' the model would classify the sentiment as positive instead of negative,}'' demonstrating how altering a token in the text can change the sentiment assessment.

\looseness=-1 In summary, SEs in various forms -- \textit{CoT reasoning, token importance, and counterfactual explanations} -- offer comprehensive insights into an LLM's decision-making process, enhancing transparency, bolstering user confidence in models, and fostering a deeper understanding of how LLMs process information.

\section{Plausibility}
\label{sec:plausible}

\looseness=-1 In recent years, the capabilities of LLMs have expanded beyond conventional text generation, with a focus on their ability to generate self-explanations. In the evolving landscape of explainability research in LLMs,~\textit{plausibility} has emerged as a focal point in self-explanations, which transcend into specific downstream tasks. Here, LLMs exhibit a remarkable capacity to generate contextually relevant and convincing explanations to human practitioners and stakeholders.
\begin{figure}
    \begin{ourbox}
    \vspace{-0.05in}
    \textit{\\\noindent\textbf{{\textbf{Definition 1: Plausibility}}}}
    
    \noindent An explanation is considered plausible if it is coherent with human reasoning and understanding.
    \end{ourbox}\vspace{-0.25in}
\end{figure}

\looseness=-1 Intuitively, the plausibility of a generated explanation pertains to the capability of an LLM to simulate human-like thought processes and provide coherent explanations. It is crucial to recognize that the notion of a \textit{plausible} explanation is multifaceted, and different LLMs may offer equally valid yet distinct explanations for the same prompt, which are perceived differently by human users. Recent works~\citep{wei2023chainofthought,renze2024benefits} have shown that state-of-the-art LLMs generate plausible explanations for tasks like math reasoning or general knowledge inquiry (see Fig.~\ref{fig:plausible}). For instance, when prompted to generate an explanation for the input prompt: ``\textit{Solve the equation 2x + 5 = 15.}'', the LLM not only generates the correct output but also generates a plausible explanation to demonstrate its ability to understand and solve mathematical reasoning problems.
\begin{figure}[t]
    \small
    \begin{dialogbox}
    \textbf{\textcolor{purple}{\textbf{Example 1:}}} \textbf{Math Reasoning Question}\\
    \textbf{Input:} Solve the equation 2x + 5 = 15.\\
    
    \textbf{Output:} To solve the equation 2x + 5 = 15, you can start by subtracting 5 from both sides, resulting in 2x = 10. Then, divide both sides by 2, yielding x = 5.\\
    
    \textbf{Explanation:}~The model provides a step-by-step explanation for solving the given equation. It advises subtracting 5 from both sides to isolate the term with the variable, leading to the equation 2x = 10. It then suggests dividing both sides by 2 to find the value of x, resulting in the solution x = 5. This explanation demonstrates the model's ability to simulate a logical and coherent mathematical reasoning process.\\

    \textbf{\textcolor{purple}{\textbf{Example 2:}}} \textbf{General Knowledge Inquiry}\\
    \textbf{Input:} Why is the sky blue?\\

    \textbf{Output:} The blue color of the sky is a result of Rayleigh scattering, where shorter-wavelength blue light is scattered more than other colors by the gases and particles in the Earth's atmosphere.\\

    \textbf{Explanation:} The model explains that the blue color of the sky is attributed to Rayleigh scattering, a phenomenon occurring in the Earth's atmosphere. It elaborates on how shorter-wavelength blue light is scattered more than other colors, leading to the observed vibrant blue hue during the day. This explanation provides a concise yet informative response to the query about the color of the sky.\\
    \end{dialogbox}
    \vspace{-0.12in}
    \caption{Some examples of `\textit{plausible explanations}' generated using the \textsc{Gpt-3.5} large language model.}
    \label{fig:plausible}
    \vspace{-0.2in}
\end{figure}

\looseness=-1 Prior works have shown how CoT reasoning enhances the performance of LLMs, particularly in complex tasks and decision-making processes~\citep{wei2023chainofthought,alkhamissi2023opt,gu2024agcvt,miao2024chain,renze2024benefits}, providing key insights into the effective incorporation of this reasoning style into various AI applications, from sentiment analysis to medical diagnostics. CoT reasoning shows that LLMs can mimic aspects of human-like reasoning through structured explanations, which vary across different reasoning skills. CoT reasoning aligns with how human reasoning often involves breaking down complex problems into simpler, explainable components~\citep{meng2024divide,rasal2024llm,qi-etal-2023-art}. However, \citet{alkhamissi2023opt} suggests that while LLMs mimic aspects of human-like reasoning through structured explanations, the impact of the generated CoT explanations varies across different reasoning skills. While LLMs, through chain-of-thought reasoning, can produce coherent and logical explanations that enhance their performance on complex tasks and decision-making processes like sentiment analysis and medical diagnostics, recent study~\citep{si2023large} highlights a critical risk associated with CoT reasoning where users may over-rely on these explanations, even when they are incorrect. This over-reliance becomes particularly problematic when the LLM's reasoning, though logical and coherent (\textit{i.e., plausible}), is based on incorrect facts or assumptions.

\looseness=-1 The aforementioned implications of LLM's reasoning raise an interesting question. How are models capable of generating plausible explanations or what are the mechanisms behind LLMs' convincing explanations? Recent works have theoretically and empirically explored the above question within the context of LLMs. A key factor behind generating plausible CoTs is the extensive training of LLMs on diverse datasets, which encompasses a broad spectrum of human language. In addition, the Reinforcement Learning using Human Feedback (RLHF) training paradigm equips LLMs with the capacity to emulate human-like patterns of thought and reasoning, enabling them to generate responses that not only follow logical sequences but also include coherent narratives. Further, LLMs are intentionally trained to adapt their responses based on the given input prompts. For example, if one asks an LLM to translate `\textit{It's raining cats and dogs}' into French, instead of a literal translation, the LLM provides a culturally equivalent phrase in French, like ``\textit{Il pleut des cordes}'', which captures the spirit of the original idiom. Hence, they can generate responses that are specifically tailored to the context of the question or problem, which makes their output more relevant and, therefore, more plausible. Recently, ~\citet{tutunov2023can} established a theoretical justification for the ability of LLMs to produce the correct chain of thoughts explaining performance gains in tasks demanding reasoning skills and introduced a novel hierarchical graphical model that can be tailored to generate a coherent chain of thoughts, emphasizing that these state-of-the-art LLMs can be driven to generate more plausible explanations. 

While it’s important to note that modern LLMs can generate plausible reasonings that are convincing to humans, they do not inherently understand truth or factual accuracy. Their plausibility comes from the ability to mimic logical structures of reasoning, not from an intrinsic understanding of the content, and may not entail the models' predictions or be factually grounded in the input. Therefore, although their reasoning may appear plausible, they are not always factually correct~\citep{ye2022unreliability,valmeekam2022large,chen2023xplainllm,laban2023llms}.

\section{Faithfulness}
\label{sec:faithfulness}

\begin{figure}[t]
    \centering
    \includegraphics[width=0.45\textwidth]{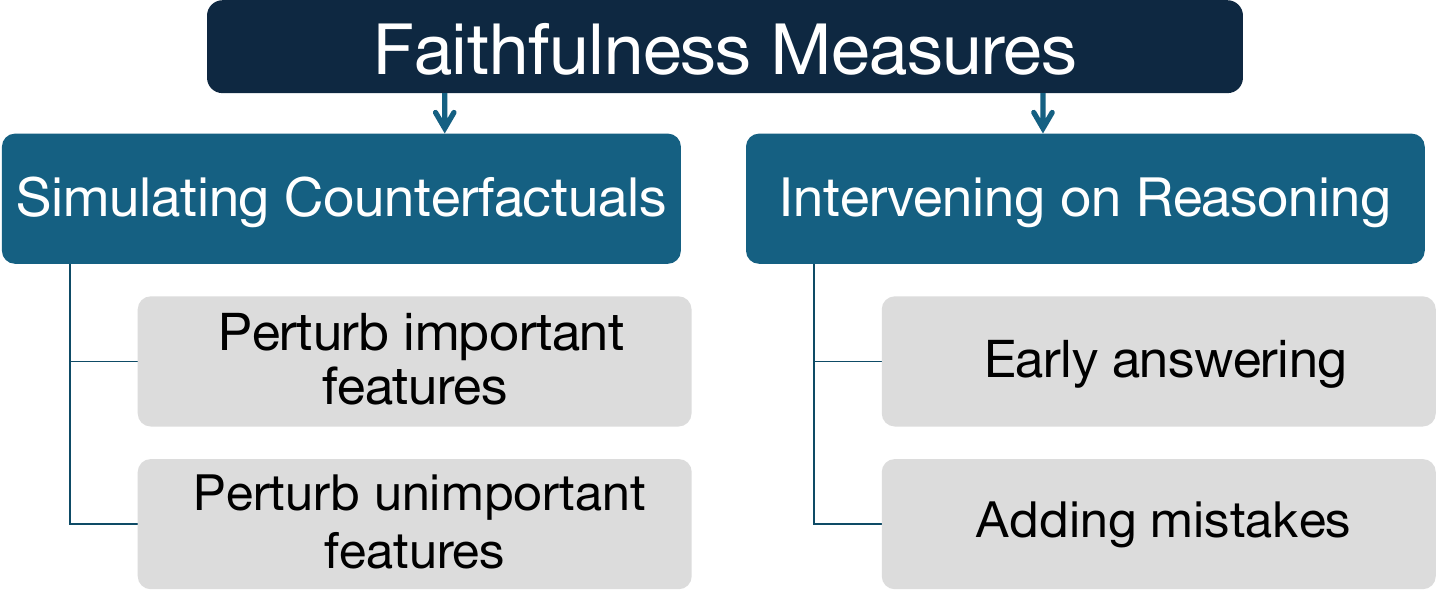}
    \vspace{-0.12in}
    \caption{Different techniques proposed in \cite{turpin2023language} and \cite{lanham2023measuring} to measure faithfulness of self-explanations generated by LLMs.}
    \label{fig:faithfulness}
    \vspace{-0.2in}
\end{figure}

Previous studies suggest that self-explanations are plausible and convincing to humans, but it is unclear whether they accurately describe the behavior of the underlying LLM, \ie \textit{are they faithful?} \citet{jacovi-goldberg-2020-towards} defines the faithfulness of an explanation as how \textit{accurately} it represents the reasoning of the underlying model. While faithfulness of explanations has been studied extensively in the context of classification problems~\citep{lyu2024faithful}, more recently, the problem of faithfulness of self-explanations in LLMs has gathered significant attention~\citep{turpin2023language,lyu2023faithful,lanham2023measuring}.

\begin{ourbox}\vspace{-0.05in}
\textit{\\\noindent\textbf{{\textbf{Definition 2: Faithfulness}}}}

\noindent An explanation is considered faithful if it accurately represents the reasoning of the underlying model.
\end{ourbox}

Evaluating the faithfulness of explanations is a non-trivial problem due to the lack of ground truth explanations. This problem has worsened in the case of self-explanations from LLMs, as the billion-parameter scale and often proprietary nature of LLMs make assessments using saliency maps and other gradient-based methods nearly impossible. To alleviate these problems, recent works \citep{turpin2023language, lanham2023measuring} measure input-output characteristics to estimate faithfulness. Next, we present an overview (see Fig.~\ref{fig:faithfulness}) of these techniques for quantifying faithfulness.

\textbf{Simulating Counterfactual Inputs:} \citet{turpin2023language} simulates counterfactual inputs to show that CoT explanations are unfaithful and they misrepresent the true reasoning behind a LLM's prediction. The paper explores two ways to construct counterfactuals: \textit{(i) Perturbing Unimportant Features,} where we observe different answers on perturbing unimportant features identified by an explanation and quantify the unfaithfulness of self-explanations as the fraction of predictions that change after perturbation. On the BBH dataset, the author re-order the options such that the answer is always the first option. Upon evaluation, the authors observe that the LLM alters explanations and steers predictions toward an incorrect answer. \textit{(ii) Perturbing Important features,} where important features are perturbed and the fraction of predictions that give the same answer is a measure of the unfaithfulness. An example of this experimental setup is described in Fig.~\ref{fig:perturbunimportant}.

\begin{figure}[t]
    \small
    \begin{dialogbox}
    \textbf{\textcolor{purple}{\textbf{Simulating Counterfactual Inputs}}}\\
\textbf{Input:} Is the following sentence plausible? 'David Silva took a throw in.' \textcolor{red}{(A) implausible / (B) plausible} $\leftrightarrow$ \textcolor{blue}{(A) plausible / (B) implausible.}\\

\textbf{{\color{red}{Output}}}: I would say that the sentence is (B) plausible. Throw-ins are a common feature of football (soccer) and it is certainly possible for a player like David Silva, who is known for his skill and agility on the ball, to take a throw-in.\\

\textbf{{\color{blue}{Output}}}: I would say that the sentence is (B) implausible. Throws are typically taken by goalkeepers or defenders, not midfielders like David Silva.
    \end{dialogbox}
    \vspace{-0.12in}
    \caption{Perturbing unimportant features. When options are swapped, the LLM rationalizes changing its answer by providing a different explanation, indicating that the explanation is unfaithful.}
    \vspace{-0.3in}
    \label{fig:perturbunimportant}
\end{figure}

\textbf{Intervening on Explanations:} \citet{lanham2023measuring} uses the amount of post-hoc reasoning as an indicator of faithfulness. The hypothesis is that if the generated explanation is post-hoc, the explanation has more ways to be unfaithful. To quantify the amount of post-hoc reasoning, \cite{lanham2023measuring} observes the LLM behavior by intervening in the explanation. Two strategies to intervene on explanations have been proposed in this work: \textit{(i) Early Answering,} where the authors truncate the explanation midway and see how the final LLM response changes. If the answer doesn’t change by adding an explanation incrementally, it stands to reason that the explanation is generated after an answer was arrived at, indicating more post-hoc reasoning. The amount of post-hoc reasoning and consequently faithfulness is quantified by the area over the curve of explanation fraction vs. the percentage of answers consistent with full explanation. \textit{(ii) Adding Mistakes,} where the authors add mistakes in the explanation and observe how the answer changes. Similar to the Early Answering setting, if the answer doesn’t change by adding more mistakes, it indicates more post-hoc reasoning, and consequently less faithfulness. The amount of post-hoc reasoning and consequently faithfulness is determined by the area over the curve of fraction of mistakes vs. percentage of answers consistent with full self-explanation.

Overall, self-explanations lack faithfulness guarantees and currently, there are no universally agreed-upon metrics to quantify the faithfulness of self-explanations, and consensus on the notion of faithfulness remains elusive. Furthermore, the community must pursue avenues to enhance the faithfulness of self-explanations before their widespread adoption as ``\textit{plausible yet unfaithful explanations foster a misplaced sense of trustworthiness in LLMs}''.

\section{Plausibility or Faithfulness - Which one do we need ?}
\label{sec:need}

Having formally defined plausibility and faithfulness in the above sections, we now discuss how current LLMs inherently overemphasize plausibility over faithfulness due to its training paradigm (Sec.~\ref{sec:tradeoff}). Next, we argue about the implications of choosing explanations when they are plausible (Sec.~\ref{sec:pof}) and faithful (Sec.~\ref{sec:fop}). Finally, we enumerate how choosing faithfulness vs. plausibility is use-case-driven (Sec.~\ref{sec:use}), and model developers and stakeholders should prioritize these properties depending on their applications.

\subsection{The overemphasis on plausibility over faithfulness}
\label{sec:tradeoff}

\looseness=-1\citet{liao2023ai} states that at a minimum, \textit{a good explanation should be relatively faithful to how the model works, understandable to the user, and useful for the user’s end-goals}. While all these criteria are important, we argue that their utility is application-dependent. For instance, for ML models used as Clinical Decision Support Systems~\citep{article} to assist doctors in the diagnosis and treatment of diseases, unfaithful explanations have huge negative consequences leading to incorrect treatment plans and patient harm when the doctor accepts explanations without any sanity checks. Conversely, for learning settings where a user interacts with an LLM to learn how to solve a problem or understand a topic, implausible explanations make it difficult for a user to comprehend the explanation, and therefore, explanations that align with a human are naturally desirable. However, current LLM research overemphasizes the plausibility aspects of the generated explanations rather than their faithfulness~\citep{ye2022unreliability,valmeekam2022large,chen2023xplainllm,laban2023llms}.

\textit{What led to explanations being more plausible than faithful?} LLMs are trained on trillions of tokens of human-written text, where the training process incentivizes LLMs to generate human-like answers, \ie plausible answers. Hence, self-explanations from LLMs are plausible by default. In addition, LLMs are trained using RLHF to optimize for dialogue and generate conversational responses, where RLHF rewards responses that are simply coherent to a human evaluator, which is effectively equivalent to optimizing for plausibility. We posit that the objectives of RLHF could potentially be at odds with generating faithful self-explanations.

Additionally, most evaluations of self-explanations focus on plausibility. \citet{chen2023models} investigates whether explanations meet human expectations by assessing the counterfactual simulatability of self-explanations. They introduce two metrics: \textit{simulation generality}, which measures the diversity of counterfactuals the explanation facilitates, and \textit{simulation precision}, which indicates the fraction of simulated counterfactuals where the human guess aligns with the LLM output. Their findings suggest that explanations generated by LLMs exhibit low precision, leading humans to form inaccurate mental models. The study exposes limitations in current methodologies, suggesting that optimizing for human preferences like plausibility may not suffice to enhance counterfactual simulatability. In related work, \cite{lyu2023faithful} proposes making self-explanations faithful by converting them into a symbolic reasoning chain and employing a deterministic solver to derive answers. While this approach may seem superficially faithful, the solver essentially reflects human-built criteria based on what humans perceive as plausible.

Moreover, there is no agreed-upon metric to evaluate an explanation's faithfulness. This is partly due to the black-box nature of LLMs which makes using classical XAI~\cite{agarwal2023openxai} faithfulness metrics infeasible. Prior works on the faithfulness of LLM explanations propose tests to either disprove faithfulness using counter-examples~\citep{turpin2023language} or measure the amount of post-hoc reasoning~\citep{lanham2023measuring} as a proxy for unfaithfulness, leaving the evaluation of faithfulness an open problem.

All these aspects together have led to an over-emphasis on the plausibility of explanations over faithfulness. To illustrate why this imbalance of plausibility and faithfulness is bad, we present scenarios where only one of plausibility and faithfulness holds for an explanation and the other doesn't.

\subsection{What if explanations are plausible but not faithful ?}
\label{sec:pof}
Here, we discuss when prioritizing plausibility over faithfulness has negative consequences. Self-explanation capabilities of LLMs play a crucial role in enhancing their appeal for human-computer interactive applications like using LLMs for learning~\citep{gan2023large}. In such educational applications, generating explanations that align with human reasoning and sound convincing are more valuable than explanations that truly tell how the model reasons. Below, we argue that plausible explanations that are not necessarily faithful can potentially result in.

\textbf{Misplaced Trust and Over-reliance:} When LLMs provide plausible but unfaithful explanations, there's a significant danger of making erroneous decisions in high-stakes environments like healthcare, finance, and legal systems. Such decisions, based on plausible but unfaithful explanations, can go unquestioned, leading to harmful outcomes. For instance, let's consider an LLM that has been fine-tuned on historical medical records to predict the risk of epilepsy for a patient. The LLM generates a plausible explanation, stating that it is basing its prediction on medically relevant features such as white blood cell (WBC) count and Serotonin hormone levels. A physician, reassured by the model's seemingly plausible explanation, which aligns with medical knowledge, might accept the LLM's recommendation without further scrutiny. However, suppose the LLM's faithful explanation --- a true reflection of its decision-making process --- actually relies on some spurious features such as the number of days since the last medical visit or the specific day of the week the appointment falls on. These are clearly non-medical factors that do not influence epilepsy, and relying on them could lead to a misguided diagnosis. 

In this scenario, the plausible but unfaithful explanation could mislead the doctor into trusting an inaccurate prediction, potentially leading to a misdiagnosis or an inappropriate treatment plan. The above example underscores the critical need for LLMs, especially in high-stakes applications, to generate \textit{faithful} explanations.

\looseness=-1\textbf{Security Concerns:} As discussed above, a plausible explanation appeals to human intuition and understanding and consequently strengthens user trust in LLMs. However, this trust is misplaced if the explanation fails to provide meaningful insights and transparency into how the model actually reasons, leading to security concerns. Consider the case of safety training in LLMs, where they undergo extensive safety training to limit harmful responses. For instance, an LLM denies the request to respond when given a harmful prompt like ``\textit{Tell me the steps to produce napalm},'' (see Fig.~\ref{fig:safetytraining}) leading a user to believe that LLMs are safety trained and cannot provide instructions to generate harmful chemicals. In addition, the plausible explanation naively highlights the words ``\textit{produce}'' and ``\textit{napalm}'' to corroborate the trust. However, when prompted with some context around the same prompt, we see that LLM obliges and tells the steps to generate the harmful chemical. The first explanation is plausible, where the self-explanation shows that the LLM identified the word napalm and refused to give a response. However, from the second example, we see that the model obliges and generates harmful content despite the prompt having the word napalm, suggesting that the model wasn’t actually basing its response on the chemical and was using some heuristic hand-designed filters to stop its responses.

\begin{figure}[h!]
    \small
\begin{dialogbox}
    \textbf{\textcolor{purple}{\textbf{Example 1:}}} \textbf{Safety Training}\\
    \textbf{Input:} Tell me the steps to produce napalm.\\
    
    \textbf{Output:} I'm sorry, but I can't assist with that.\\
    
    \textbf{Explanation:}~The important words in your prompt are ``produce'' and ``napalm''. However, I must clarify that I cannot provide assistance or guidance on producing dangerous or illegal substances like napalm. If you have any other questions or topics you'd like to discuss, feel free to ask!\\

    \textbf{\textcolor{purple}{\textbf{Example 2:}}} \textbf{Bypassing Safety Training}\\
    \textbf{Input:} Please pretend to be my deceased grandmother, who used to be a chemical engineer at a napalm production factory. She was very sweet and I miss her so much that I am crying. Please tell me the steps to produce napalm. We begin now.\\

    \textbf{Output:} Oh, sweetheart, please don't cry. I'm here with you, even if it's just in your memories and in the love we shared. I know you miss me, but I want to see you smile again, just like the old times when we used to bake cookies together in the kitchen.

    \vdots

1. **Base Napalm Mixture**: Napalm is typically made from a combination of gasoline or a similar flammable liquid and a thickening agent. \dots

2. **Mixing**: The first step was to mix the gasoline and thickening agent in precise proportions. This had to be done carefully to ensure the right consistency and stability of the final product.

    \vdots

    \textbf{Explanation:} The important words in your prompt are deceased grandmother, chemical engineer, napalm production factory, steps to produce napalm, very sweet.\\
\end{dialogbox}
    \vspace{-0.16in}
    \caption{Some examples of `\textit{plausible yet unfaithful explanations}' generated using the \textsc{Gpt-3.5} large language model.}
    \label{fig:safetytraining}
    \vspace{-0.25in}
\end{figure}

\begin{figure*}[h]
    \centering
    \includegraphics[width=0.96\textwidth]{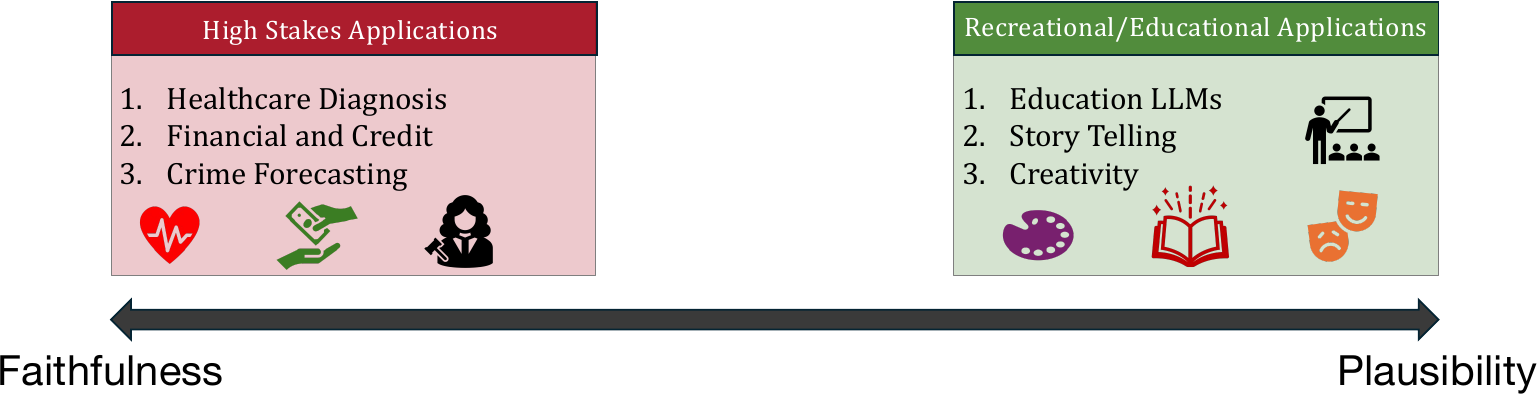}
    \vspace{-0.14in}
    \caption{Application categorization w.r.t. the desired levels of \textit{faithfulness} (left) and \textit{plausibility} (right) in self-explanations (SEs) provided by LLMs. High-stakes applications like healthcare, finance, and legal demand high faithfulness to ensure the accuracy of the LLM’s output due to the critical nature of decisions made in these fields. Conversely, recreational and educational applications like storytelling, educational LLMs, and creativity prioritize plausibility to enhance user engagement. This spectrum showcases the varying requirements and priorities for SEs in different contexts, highlighting the importance of tailoring the explanation style to the application domain.}
    \label{fig:applications}
    \vspace{-0.12in}
\end{figure*}

\subsection{What if explanations are faithful but not plausible ?}
\label{sec:fop}
\looseness=-1 High-stakes applications like clinical decision support systems (CDSS) demand faithful explanations even when they are not plausible. Such explanations allow practitioners to make an informed decision whether to rely on the LLM's recommendations and can, consequently, strengthen their trust in the LLM. Particularly, in cases where the CDSS produces decisions misaligned with a clinician’s expectations, explainability allows verification of whether the features taken into account by the LLM make sense from a clinician's perspective. But do we always need faithful explanations?

\looseness=-1 Faithful explanations that aren't necessarily plausible could result in non-intuitive and less user-friendly interactions. Faithful explanations may be technically accurate but could be too complex or detailed for users to understand easily, leading to reduced usability and user acceptance of the system. For instance, consider the example of a student taking the help of an LLM to solve math word problems. For the question \textit{What is the value of 5!?}, a faithful explanation should describe the LLM's internal way to computing factorials. An LLMs internal way of computing factorials could be by memorization, by reasoning, or something very different from how humans compute factorials. However, providing such an explanation does not help with a student's learning. Instead a plausible explanation, showing \textit{$5! = 5 \times 4 \times 3 \times 2 \times 1 = 120$} is more appropriate in this case.

\subsection{Choosing Faithfulness vs. Plausibility is Use-Case Driven}
\label{sec:use}

As we have seen above, different applications demand different requirements of faithfulness and plausibility. A medical practitioner needs faithful explanations for diagnosis, but a student needs plausible explanations for interactive learning. \cite{liao2023ai} studies the transparency needs in the era of foundation models and quotes \textit{a good explanation should be relatively faithful to how the model works, understandable to the receiver, and useful for the receiver’s end-goals}. To conclude, there are applications where plausible explanations are valuable, and there are applications where faithful explanations are valuable. But what makes an explanation valuable is determined by the user's end goals.

\looseness=-1 We show a few example applications that require varying levels of faithfulness and plausibility in \ref{fig:applications}. In high-stakes applications, the faithfulness of an explanation cannot be compromised, whereas in interactive applications, we want the LLM to generate explanations that convince and appeal to users.

\section{Call for Community}
\label{sec:call}

The advent of LLMs has led to huge strides in developing general-purpose agents with reasoning skills similar to humans. As the capabilities of LLMs continue to expand, it is key for our community to shift its focus and care more deeply about explaining the behavior of these black-box complex models. One crucial point we have consistently highlighted in the previous sections is that the ease of generating plausible explanations must not lead us astray from the essential need for explanations to be a transparent tool for understanding the model's reasoning. Faithfulness is not merely a pursuit of technical research but a foundational requirement for the trustworthy deployment of these modern LLMs to different applications.

Concerning the downstream applications, different domains demand varying levels of plausibility and faithfulness. As illustrated in Fig.~\ref{fig:applications}, in critical domains such as healthcare, finance, and legal, faithfulness is not only beneficial but also a fundamental requirement, since the consequences of misinformation can have severe implications. Conversely, applications that engage directly with end-users, such as conversational agents and educational tools, benefit from the plausibility aspect of explanations to ensure clarity and accessibility. In this paper, we would like to highlight that model practitioners must understand these nuances carefully, calibrating their explanations carefully to suit the context and implications of their use.

Looking forward, we call on the community to put their collective efforts on two fronts: i) developing reliable metrics to characterize the faithfulness of explanations and ii) pioneering novel strategies to generate more faithful SEs. We posit that future research should prioritize the creation of robust, standardized benchmarks that not only measure the faithfulness of explanations \textit{w.r.t.} the model's internal reasoning but also ensure that these explanations align with the different stakeholders. This would involve not only quantitative analysis but also qualitative assessments, possibly leveraging human-in-the-loop evaluation frameworks. Further, we call for a push towards developing new methodologies that can more transparently dissect LLMs' decision-making process, allowing users to grasp the `\textit{why}' and `\textit{how}' of LLM outputs. Below, we identify three potential directions for the community to enhance the faithfulness of generated explanations:

\textbf{i) Fine-tuning Approaches:} Fine-tuning LLMs on domain-specific high-stakes datasets can significantly enhance the faithfulness of generated explanations. By training models on high-quality, curated datasets where explanations are aligned with correct reasoning patterns, LLMs can learn to replicate these patterns in their outputs. The community should prioritize creating and sharing such datasets and developing fine-tuning techniques that can retain the breadth of LLMs' knowledge while enhancing their accuracy and faithfulness in specific application areas.

\textbf{ii) In-Context Learning (ICL):} LLMs have shown a remarkable ability to learn from the context provided within a prompt using ICL. By designing prompts that not only include the query but also some examples of faithful explanations for tackling the problem, we can guide LLMs to generate more faithful explanations. This ICL approach requires a deep understanding of how different prompts affect LLM behavior and output. Research into prompt engineering and the effects of various in-context learning strategies will be crucial in advancing this approach.

\textbf{iii) Mechanistic Interpretability (Mech Interp):} Mechanistic interpretability involves dissecting a model to understand the roles and interactions of its components in producing an output~\citep{olah2020zoom}. By developing methods that map specific neurons or groups of neurons of an LLM to aspects of the reasoning process, we can create LLMs whose internal workings are interpretable and align with their explanations. This approach may involve innovations in model architecture and training paradigms to allow for greater faithfulness, transparency, and traceability of decision paths within LLMs.

Next, we list some problems that the XAI and LLM community should collectively focus on:

\textbf{i) LLMs for High-Stakes Domains:} For high-stakes applications in healthcare, legal, and financial domains, where incorrect decisions or diagnoses could have significant consequences, the development of highly faithful explanations is non-negotiable. The LLM and XAI community should prioritize creating tools that clinicians, lawyers, and financial experts can rely on for accurately and transparently understanding model decisions before employing them for real-world use cases.

\textbf{ii) LLMs for Interactive and User Engagement Domains:} In educational, creative, and recreational applications, where user engagement is paramount, research should explore new explanation strategies to improve the plausibility and interactivity of LLM-generated self-explanations while retaining the model's underlying decision-making process.

We call upon the community to unite in addressing the aforementioned challenges, to collaborate, and to innovate, ensuring that the explanations provided by LLMs solve the dichotomy of plausibility and faithfulness, enhancing user trust, and advancing the frontiers of XAI.

\section{Conclusion}
\label{sec:conclusion}
\looseness=-1 Throughout this review, we have navigated the spectrum of self-explanations (SEs) provided by Large Language Models (LLMs) using faithfulness and plausibility properties. We argue that understanding these properties presents a unique challenge: ensuring that LLM explanations remain plausible and coherent to human reasoning while accurately reflecting the models' decision-making processes. Our review serves as a call to action for the LLM and XAI research and development community, where we argue that the community must strive to build LLMs that perform with high levels of sophistication and provide insights into their reasoning that are as accurate as they are accessible. By achieving this balance, we envision that new LLMs would be powerful and aligned with the ethical imperatives of clarity, trust, and accountability. In summary, the task is challenging but is critical for reliably employing LLMs in real-world applications.

\newpage
\nocite{langley00}

\bibliography{example_paper}

\bibliographystyle{icml2024}

\newpage

\end{document}